# CONCEPTUAL MODELLING AND THE QUALITY OF ONTOLOGIES: ENDURANTISM VS. PERDURANTISM

Mutaz M. Al-Debei[1], Mohammad Mourhaf Al Asswad[2], Sergio de Cesar[3] and Mark Lycett[4]

[1]Department of Management Information Systems,
The University of Jordan, Amman, Jordan
m.aldebei@ju.edu.jo
[2]Department of Information Systems and Computing,
Brunel University – West London, London, UK
{mohammad.alasswad,Sergio.deCesare,mark.lycett}@brunel.ac.uk

## ABSTRACT

*Ontologies are key enablers for sharing precise and machine-understandable semantics among different applications and parties. Yet, for ontologies to meet these expectations, their quality must be of a good standard. The quality of an ontology is strongly based on the design method employed. This paper addresses the design problems related to the modelling of ontologies, with specific concentration on the issues related to the quality of the conceptualisations produced. The paper aims to demonstrate the impact of the modelling paradigm adopted on the quality of ontological models and, consequently, the potential impact that such a decision can have in relation to the development of software applications. To this aim, an ontology that is conceptualised based on the Object-Role Modelling (ORM) approach (a representative of endurantism) is re-engineered into a one modelled on the basis of the Object Paradigm (OP) (a representative of perdurantism). Next, the two ontologies are analytically compared using the specified criteria. The conducted comparison highlights that using the OP for ontology conceptualisation can provide more expressive, reusable, objective and temporal ontologies than those conceptualised on the basis of the ORM approach.*

## 1. INTRODUCTION

Ontology research has attracted increasing attention in Information Systems (IS) design and development (Wand and Weber 2002; Fonseca 2007). In this context, ontologies are recognised as a useful means for achieving semantic interoperability between different systems. This is because ontologies can capture semantics to facilitate shared understanding between different parties (Ouksel and Sheth 1999). The weight of ontologies however comes from the fact that they are considered important backbones for many organisational applications in areas including, but not limited to, knowledge engineering, information integration and software development. In addition, they are highly significant for the Semantic Web and Semantic Web services (AL Asswad et al. 2009). Hence, information systems that make use of explicit and formally defined ontologies have been described as ontology-driven systems (Guarino 1998). Such ontologies are also referred to as IS ontologies (e.g., Smith, 2003) or computational ontologies (e.g., Kishore et al., 2004).

Simply put, an ontology is an "engineering artefact" (Guarino 1998) representing a particular phenomenon or domain of knowledge. Ontologies are generally composed of concepts, relations between these concepts and axioms to restrict the interpretation of concepts and are (ideally) precise, machine-understandable and signify shared representations of real world phenomena. Consequently, it is important that ontologies are of a good quality, in order that they serve their intended purposes and be shared as well as reused by different applications (Guarino 2004). The quality of ontological models can be evaluated based on the models' semantic preciseness and richness; that is, the extent to which the ontology is describing a particular phenomenon abstractly, but accurately and meaningfully. The quality of ontologies is heavily based on the adopted engineering methodology (Nicola et al. 2009) in which conceptualisation (i.e., creating conceptual models) is a significant activity (Uschold and King 1995; Al-Debei and Fitzgerald, 2009). Conceptual modelling, however, can be defined as "the activity of formally describing some aspects of the physical and social world around us for the purposes of understanding and communication" (Mylopoulos 1992 p. 389).

In this paper, we place an emphasis on conceptual modelling paradigms and their impact on the quality of the developed ontologies. In line with Jarrar et al. (2003), we argue that developing a qualified ontology is strongly based on the conceptual modelling paradigm employed. There are two reasons for this position: (a) conceptual modelling can be seen as a tool to analyse the structure of a given reality (Guarino 1998); and (b) paradigm modelling constructs are utilised to represent the required ontology. For example, Jarrar et al. (2003) argue that the decision to model a concept as a class or a property is based on the employed conceptual modelling paradigm. Therefore, extra care must be taken in order to clearly define objects' identities in the sense of clarifying their distinctive features. In other words, the more issues about objects (e.g., relations and instances) considered during conceptual modelling, the more the potential for a richer and more accurate conceptualisation (Spyns et al. 2002).

Modelling paradigms can generally be classified as three-dimensional (i.e., endurantism; for example, Object-Role Modelling (ORM) and Object-Orientation) and four-dimensional approaches (i.e., perdurantism; for example, the Object Paradigm (OP)). These paradigms perceive real-world phenomena from different philosophical standpoints, thus allowing different conceptual models to be produced. This paper aims to demonstrate the quality variations in ontologies across the two modelling paradigms (three-dimensional vs. four-dimensional), by utilising the ORM and OP approaches as representative cases. To this aim, a bookstore ontology conceptualised on the basis of ORM is re-engineered into an ontology modelled according to OP to highlight differences.

The remainder of this paper is structured as follows. Section 2 provides a theoretical background relating to ontology, ontology engineering and conceptual modelling paradigms. Section 3 explains the research design. Section 4 demonstrates the re-engineering process and provides a comparison between the models to show advantages and deficiencies of the two models. Section 5 discusses the significant implications for both theory and practice. Finally the conclusions are presented.

## 2. THEORETICAL BACKGROUND

### 2.1 Ontology and Ontology Engineering

Ontology is a term that originated in philosophy and refers to the systematic explanation and study of the nature of existence, or being (Chandrasekaran et al. 1999). The term has been subsequently borrowed by the information systems and computing disciplines (e.g., Wand and Weber 1990; Guarino and Welty 2002) and changed somewhat. For example, Gruber (1995) argues that philosophical ontology limits the ontological representation to class definitions and taxonomies; thus more constructs such as axioms are required to constrain the interpretation of defined concepts.

Basically, ontology is a constructed model for a particular domain representing a real-world phenomenon. In computational terms, an ontology is most commonly defined as a formal explicit specification of a shared conceptualisation (Gruber 1993). The inclusion of the terms 'explicit' and 'conceptualisation' in this definition is highly significant. The term 'explicit' highlights knowledge externalisation as one of the main characteristics and reasons for ontology developments; whilst 'conceptualisation' is a key attribute of an ontology. Fundamentally, conceptualisation is what makes ontologies sharable as it refers to the meanings captured through concepts and not the terms themselves. Furthermore, conceptualisation implies abstraction which signifies that an ontology represents only knowledge regarded as core in the specific domain. So, in practice, computational ontology provides a definition of concepts, axioms, and their properties in a formal, precise and shared format (Jasper and Uschold 1999; Kishore et al., 2004).

Ontology engineering is a subfield that covers issues related to ontology development and use throughout its life span (Gomez-Perez et al. 2004). Ontology development covers a set of activities conducted during conceptualisation, design, implementation and deployment phases (see Devedzic 2002). There are few approaches or methodologies for constructing ontologies however (e.g., Gruninger and Fox 1995; Fernandez-Lopez et al. 1999; Al-Debei and Fitzgerald 2009). Pinto and Martins (2004), for example, argue that the ontology engineering process is composed of the following five phases: specification, conceptualisation, formalisation, implementation, and maintenance; whilst Uschold and King (1995) report four main stages for ontology development: identifying purpose, building the ontology (ontology capture, ontology coding, and integrating existing ontologies), evaluation, and documentation. Irrespective of any differences among existing methodologies, however, they all regard conceptualisation as a major activity in ontology engineering.

Conceptualisation, as a step, does not always necessarily result in an explicit conceptual model; that is represented by a particular conceptual modelling language. This step could produce notes, comments, interpretations, and schemas that do not follow any particular modelling language. Whether the aim of conceptualisation is to end up with an explicit conceptual model or not, the adopted modelling paradigm during this step is very significant. This is because the paradigm can inform the methodological guidelines followed and determine (a) what objects shall be considered during conceptualisation; and (b) how different objects are represented in order to produce high quality ontologies. Indeed, conceptualisation, to be effective, needs to make use of significant, accurate and clear concepts and language that are easily understood by their intended users and at the same time faithfully representing real-world phenomena. In the next section, we discuss in more detail the paradigms used in conceptual modelling along with their philosophical differences.

## 2.2 Conceptual Modelling Paradigms: Endurantism vs. Perdurantism

Conceptual modelling paradigms can be categorised based on the way we perceive real world phenomena and the way in which we model changes in time. In regard to this matter, philosophers have distinguished between endurantism (also called three-dimensional or 3D) and perdurantism (also called four-dimensional or 4D) paradigms for viewing and representing diachronic identities. These two paradigms are philosophically different and in some points divergent. Thus, they represent a significant and fundamental ontological choice for ontology engineers when conceptualising real world domains. The decision in this phase is important, as it is most likely to greatly influence the quality of the ontology and software applications that may use produced ontologies at later stages.

Endurantist paradigms such as ER (Entity-Relationship), OO (Object-Oriented) and ORM (Object Role Modelling) assume that objects have three spatial dimensions and exist in full at each moment of their lifetime (Hales and Johnson 2003). Hence, endurantist objects normally have no time dimension when they are defined in an ontology, and if they have to be linked to time then they have to be indexed separately and ineffectively to points in time or intervals. This type of paradigm assumes that the same object can exist over time and thus may be fully identified at different points in time (e.g., *t1, t2, t3... tn);* at each point in time objects are viewed only from the present.

Retrospectively, endurantist approaches seem to be inconsistent with the indiscernibility of identicals, i.e., principle about the meaning of identity (Pease and Niles, 2002). For example, suppose we have a Car (i.e., object) whose colour (i.e., property) changes at a point in time (*t*). According to endurantism, Car before (*t*) is identical to Car after (*t*). But as explained in Pease and Niles (2002), the indiscernibility of identicals principle postulates that something A is identical with something B if every property that can be ascribed to A can be ascribed to B and vice versa. Therefore, we agree with Pease and Niles (2002) that a contradiction may be generalised from endurantism. With hindsight, we believe that one of the major challenges while using endurantist paradigms in conceptual modelling is the identification of "essential properties" that do not change over some period of time (Krieger et al., 2008) so as to avoid contradiction. For example, *weight* is unlikely to be considered an essential property of a person if periods of time are taken into account, if endurantists wish to eliminate any inconsistency in regards to the indiscernibility of identicals principle. However this represents a compromise (sacrificing some properties and details which are perhaps important so as to avoid contradiction) taking place during the conceptualisation phase that is most likely to affect the quality of the ontology developed at a later phase. However, endurantist conceptual modelling is widely used in areas such as database design and information systems development and have also been utilised within ontology development (Jarrar et al. 2003).

Perdurantist approaches, such as OP, assume that objects have four dimensions (spatial and temporal) and just partly exist at each time instant of their life span (Hales and Johnson 2003). That is entities only exist for some period of time and continually change over such period. Such entities are unfolding themselves over time in successive temporal parts (Semy et al., 2004). Therefore, objects are viewed from past, present, and future. According to this paradigm, entities are usually referred to as "space-time worms" or a slice of such a worm (Loux, 1980; Sider, 2001) given that they are identified based on space and time dimensions. Thus, unlike endurantist paradigms, semantics that are embedded in changes of time are considered by perdurantist paradigms as a core aspect of the approach. The main advantage of perdurantist approaches is simplicity as everything (e.g., objects and processes) is treated in a similar way as a space-time worm. Ironically, this advantage is regarded by some other researchers as a disadvantage given that in this way perdurantist approaches do not distinguish between objects and processes and thus regarded as counterintuitive (Pease and Niles, 2002). Perdurantist conceptual modelling has been proved useful in areas such as database design (Erwig et al. 1999), systems analysis and design (Parent et al. 1999), geographic information systems (Xu et al. 2006) and ontology engineering (Partridge 2005).

The differences between both of these approaches lay in their underlying philosophical foundations (see Table 1). The consideration of spatio-temporal extensions of any object in any perdurantist paradigm intrinsically represents temporality, whilst those based on endurantism do not take this extension into consideration. Indeed, endurantist approaches are pre-equipped to deal with spatial extensions only, but require fundamental changes to be able to deal effectively with temporality. Furthermore, in perdurantist approaches the identity of each object is clearly defined, which is not the case in endurantist approaches as they assume that an object is completely existent at any one point in time. For example, if two real-world objects have the same attributes and attribute values in OO (taken as an endurantist approach) then they are recognised as the same object. But this is not true for perdurantist approaches such as the OP as the objects may have different temporal extensions.

*Table 1. Differences between Endurantist and Perdurantist Approaches*

| Endurantism (3D) | Perdurantism (4D) |
|---|---|
| 1. Objects have only spatial dimensions | 1. Objects have spatial and temporal dimensions |
| 2. Objects are wholly present at any point of time during their lifetime | 2. At any given time a 4D object is only partially present |
| 3. Objects are viewed from the present. The default is that statements are true now. | 3. Objects from the past, present, and future all exist. |
| 4. Objects do not have temporal parts. | 4. Objects extend in time as well as space and have temporal parts as well as spatial parts. |
| 5. Different objects may coincide at a point in time, i.e., occupy the same 3D extension (non-extensionalism). | 5. When two objects have the same spatio-temporal extent, they are the same thing (the extensionalist criterion of identity). |
| 6. Time and space are treated separately. | 6. Time and space are unified. |
| 7. Understand change in terms of things. | 7. Understand things in terms of change. |

In explaining these differences in more detail, we present ORM as a representative of endurantist approaches and OP as an example of perdurantist paradigms and briefly explain the reasons for choosing these two approaches.

### 2.2.1 ORM (Object Role Modelling)

ORM is a conceptual paradigm for modelling and querying information systems. Natural language and diagrams are used in ORM to represent a phenomenon (Krogstie et al. 2007: p.23). ORM depicts a reality using objects (entities having values) that play some roles to participate in relationships (Halpin 1998). An ORM model can therefore be depicted as a network of entity and relationship types representing a specific domain. Entities can be lexical (i.e., utterable), such as 'colour' and 'name', or non-lexical (i.e., unutterable), such as 'car' and 'man' (Jarrar et al. 2003) – though such a classification is based only on linguistic distinctions. Role types describing the type of fact which occurs between entity types in the Universe of Discourse (UoD) are verbalised and presented on a conceptual schema diagram using established ORM graphical notations. Furthermore, the ORM approach allows for the definition of subtype and whole-part relationships. Constraints such as asymmetry, uniqueness, cardinality and intransitivity can be also specified in ORM models. To clarify the main modelling constructs in ORM, consider the following simple example: a man is driving a car. The concepts 'man' and 'car' are both considered as entity types in ORM notation. A man has a role in that he *drives* a 'car' while a 'car' has a role of being driven by a 'man'.

In contrast to other 3D modelling paradigms, ORM does not use attributes, although relationships are used to indirectly represent entity-attribute relationship. To give just one example, if we assume a 'country' entity to have a 'continent' as one of its attributes, this kind of relationship is represented in ORM as follows; 'a country is *located in* a continent' where '*located in*' is the relationship and both 'country' and 'continent' are regarded as entities.

One of the critical success factors of ORM is the availability of conceptual modelling tools that are based on its paradigm, such as Microsoft's VisioModeler, DogaModeler and Visio for Enterprise Architects (VEA). Such tools allow an automatic generation of a normalised relational database schema based on modelling a particular UoD in ORM (Jarrar, 2007). Indeed, the availability of such tools has led to the widespread use of ORM. The ability to translate ORM schemas into pseudo natural language statements is another advantage contributing to the success of this conceptual modelling approach (Krogstie et al. 2007). Such facility provides an effective means for non-computer experts and scientists to create, modify, and evaluate the knowledge needed in an information system regarding a particular real world phenomenon (Jarrar, 2007).

One can argue that ORM has some advantages over other endurantist paradigms (e.g., Halpin and Bloesch 1998; Jarrar et al. 2003; 2007) in that: (1) It enables an easy and effective definition of constraints since imposing constraints on relationships is easier and more effective than on attributes; and (2) the use of relationships only can eliminate the confusion caused when taking a decision in OO or ER to model something as a relationship or an attribute. Being an attribute-free approach also provides immunity from changes that may cause attributes to be remodelled as entity or relationship type. Moreover, dominant logic-based ontology representation languages, such as the Web Ontology Language (OWL), provide ontological constructs for modelling properties as relations only and not as attributes. Subsequently, translating ORM-based conceptual models into OWL, for example, is more natural and a less confusing process than the translation from ER or OO into OWL. We consider that using ORM as a representative for endurantist paradigms, is significant.

### 2.2.2 Object Paradigm (OP)

Having recognised the importance of the temporal dimension in modelling, proposals have been developed for more naturally incorporating this dimension into conceptual modelling. To give just a few examples: More et al. (2001) and Hadzilacos and Tryfona (1997) add temporal semantics to OO and ER respectively; Allen et al. (1995) present a spatiotemporal model for explicitly modelling temporal aspects of Geographic Information Systems (GIS); and Worboys (1994) proposes a framework for modelling spatiotemporal data using two spatial and two temporal extensions. These frameworks suffer from some or all of the following however: (1) being application-dependent and thus limited in their general usefulness; (2) lack of identification and clarification of the modelling constructs and their use; (3) being theoretical proposals that have not been tested empirically; and (4) being mathematically-based and not focused towards conceptual modelling design.

The Object Paradigm (OP) proposed by Partridge (2005) has advantages in the area of ontology engineering over other perdurantist approaches because it: (1) provides holistic modelling constructs that are suitable for ontology conceptualisation; (2) offers detailed ontological modelling guidelines through the BORO (Business Object Reference Ontology) process; and (3) creates patterns for modelling frequently occurring problems such as geographic areas and naming patterns. In order to provide clear referencing to things in the world, OP considers everything as objects where object identity is a key factor for distinguishing objects from each other. Object identity is defined as an object's spatio-temporal extension in the universe. For example, a book identity is defined by its three spatial dimensions that this particular book occupies in addition to its temporal dimension which is represented by the book's overall life span.

The ontological constructs used in the OP can be classified as: (a) individuals; (b) classes; and (c) tuples (Partridge 2005). Individuals can be depicted as four-dimensional things that persist through time. In other words, individuals are perceived as particular objects that cannot be further instantiated, such as 'John Smith' and 'Adam Smith' being individuals of a 'People' class as an example. Tuples are relations (properties) between individuals. For example, 'Adam Smith' *is son* *of* 'John Smith'. Classes are types of objects such as 'People', or even more specifically tuple classes (or tuple types). An example of a tuple class is represented by 'is son of' which represents all the 'is son of' tuples between sons and fathers. A class extension is the sum of extensions of its members.

Importantly, the OP considers the semantics of changes happening to objects. Changes in OP can be modelled through 'states' and 'events'. States are considered as temporal parts of individuals. An example is the 'father' state of the individual 'John Smith' as 'John Smith' could go through many states over time. Events are types of individuals that do not persist over time as they just happen. An example of an event is the 'birth of Adam Smith' which initialises the father state of 'John Smith'.

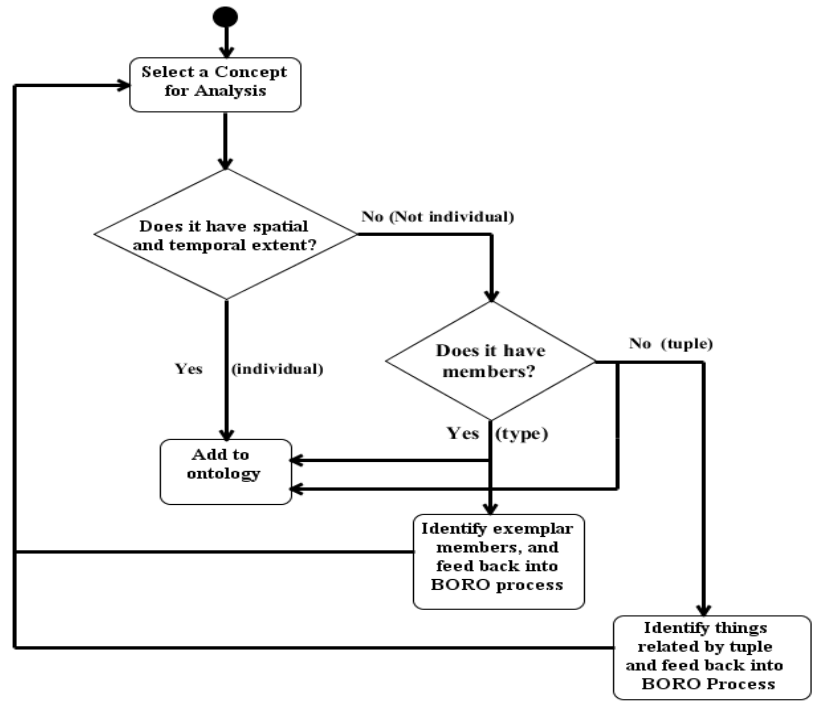

*Figure 1. BORO Process (Partridge 2001)*

Partridge (2001) developed a systematic methodology for modelling ontologies based on the OP (see Figure 1). The methodology is called BORO (Business Object Reference Ontology). It guides an ontology modeller during the process of analysing the phenomenon under consideration. This methodology is explained as follows: (a) a concept is selected for analysis; (b) if this concept has a spatial and temporal extension then it is an individual, otherwise, it is either a class or a tuple; this object is (c) a class when it can be instantiated or (d) a tuple when it cannot be instantiated.

## 3. RESEARCH DESIGN

This paper utilises a bookstore ontology that is modelled using the ORM approach. This ontology is then re-engineered according to the OP with the aim to analytically delineate the impact of the conceptual modelling paradigm on the quality of the developed ontologies. While ORM is used here to represent endurantism, the OP is used to represent perdurantism. The first issue needing to be tackled is that of answering the question relating to the definition of 'semantic quality' in the context of ontologies. To this aim, we analysed and synthesised the relevant literature (Gruber 1995; Gomez-Perez 2001; Wand and Weber 2002; Shanks et al. 2008) and established a set of criteria which are considered important pertaining to the quality of ontologies:

- **Expressiveness**, which is the ability to faithfully conceptualise the relevant details of a particular domain and represent them in an understandable and unambiguous manner.
- **Temporality**, which is the ability to track changes of ontology objects over time.
- **Extensibility**, which is the ability to expand the ontology gracefully in order to be able to cope and capture future needs.
- **Objectivity**, which is the ability to produce conceptualisations of ontologies in a smooth, managed and guided way. This criterion refers to the existence of techniques and methodologies that could guide modellers during the conceptualisation phase of ontology development in order to reduce modelling errors and confusions.

An existing and published bookstore ontology was chosen as a basis for re-engineering (see Jarrar et al. 2003). This ontology was chosen for the following reasons: (1) the ontology is modelled using one of the endurantist approaches (i.e., ORM) and thus can be re-engineered into perdurantism; (2) the use of the bookstore ontology is simple and understandable by a general audience; (3) the journal paper that the ontology is derived from is highly cited; and (4) the main focus of the selected paper is conceptual modelling for ontologies and so consistent with our main objective. The re-engineering process is led by a set of competency questions that show the deficiencies of the existing ontology. Competency questions represent a reasonably accepted technique that is useful when (re-)engineering ontologies (Gruninger and Fox 1995; Uschold and King 1995). The BORO process was followed in translating the ORM ontological model into OP ontology. Furthermore, we applied the established quality and evaluation framework to analytically compare the two ontological models showing their competencies and drawbacks. Based on this comparison exercise, the implications of this paper for theory and practice are clarified.

## 4. ILLUSTRATIVE CASE: RE-ENGINEERING A BOOKSTORE ONTOLOGY

Jarrar et al. (2003) modelled a bookstore ontology based on the ORM paradigm according to the domain knowledge presented in Table 1. The resulting ontology is shown in Figure 2. This ontology has three unutterable entity types – 'Product', 'Book' and 'Price' - and five utterable entity types – 'ISBN', 'Title', 'Author', 'Value' and 'Currency'. The decision to model an entity as an utterable (lexical) or unutterable (non-lexical) entity is merely based on a linguistic distinction. If an entity is to be linguistically considered as an identifier or descriptor for another entity then this entity is utterable, otherwise it is unutterable entity. In ORM, the identity of an unutterable entity is based on the uniqueness and mandatory properties. For instance, if an 'ISBN' is the unique and mandatory identifier for all instances of the concept 'Book', then we can use 'ISBN' as an identity property for 'Book'. This identifier is indicated by a dot alongside the oval representing the class of books in Figure 2. Roles are represented as rectangular boxes in the ontology.

*Table 2. Bookstore ontology base (Jarrar et al. 2003)*

| Term1 | Role | Term2 |
|-------|------|-------|
| Book | Is_A | Product |
| Book | Has | ISBN |
| Book | Has | Title |
| Book | WrittenBy | Author |
| Book | ValuedBy | Price |
| Price | Has | Value |
| Price | Has | Currency |

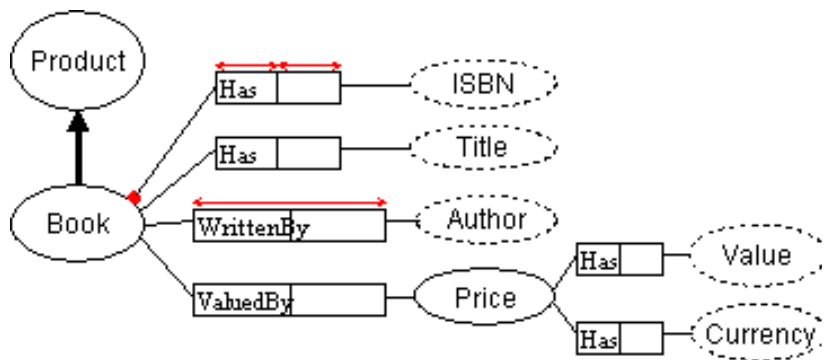

*Figure 2. ORM Bookstore ontology (Jarrar et al. 2003)*

In analysing and examining the ORM bookstore ontology using the established criteria, it is apparent that it does not faithfully express some important domain details. The key issue here is how well the ontology represents reality and whether it represents reality precisely or not, which can be examined using the following competency questions.

Q1.1 How many editions of a certain book have been published?

Q1.2 Who are the authors of a specific edition of a certain book?

Q1.3 How many copies of a certain edition have been produced?

Q1.4 What is the ISBN of a certain edition of a specific book?

The ORM ontology is unable to provide answers to the previous four questions since it does not capture the differences between editions and copies of books. Catering for these differences requires careful analysis of the dimensions of entities in this ontology. According to the OP, the examination of the temporal and spatial extensions of the book concept leads us to differentiate amongst books, book editions, and their copies. Arguably, this is more consistent with real world books since a book may have more than one edition, each with a different ISBN. A new edition of the same book could also have a different title. Moreover, a different set of people can participate in writing different editions of the same book and each edition may normally have more than one copy.

So as to capture reality more accurately, we re-engineer the ORM ontology and add both 'BookEditions' as well as 'BookCopies' to the OP ontology. Basically, the 'Books' class is modelled as a composition of its 'BookEditions' as Book editions succeed each other in a temporal sequence. 'BookCopies' are the copies of editions which are the actual products that are sold or exchanged; thus 'BookCopies' is modelled as a subclass of 'Products' (see Figure 3).

Although the ORM ontology does not model 'BookEditions' and 'BookCopies', BookEditions could be added to the ORM ontology (for example, Book has Editions where 'has' is a role). Capturing 'BookCopies', however, requires an analysis of the spatio-temporal extensions of a book in order to recognise the differences between 'Books' and 'BookCopies'. The latter can only be done using the OP paradigm because of its identity analysis mechanism and philosophy that is based on spatio-temporal extension. It is worth mentioning here that the re-engineering process follows the sequence in the BORO process discussed previously and thus more objective. This process starts by selecting a concept followed by analysing its spatial and temporal dimensions in order to figure out whether it is an individual, class or property.

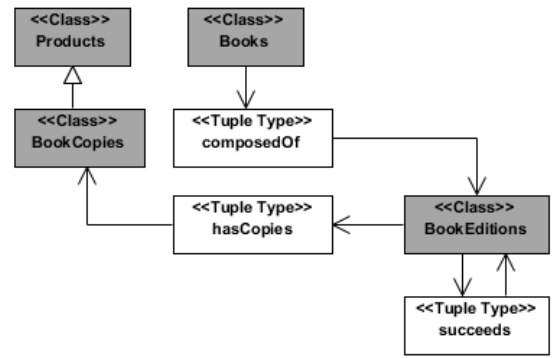

*Figure 3. Re-engineering the Book Class into Book, Book Editions, and Book Copies Classes*

Importantly, the re-engineering conducted so far allows us to achieve *clarity* as one of the most significant evaluation criteria of ontologies that enhances their expressiveness. Indeed, an ontology needs to successfully and objectively communicate the intended meaning of defined terms (Gruber 1995). Furthermore, the clarity and validity of the ontological expressiveness require the absence of the following deficiencies (Wand and Weber, 2002; Shanks et al., 2008):

- **Construct overload**, where two or more ontological constructs map to one modelling (i.e., grammatical) construct.
- **Construct redundancy**, where two or more modelling constructs map to one ontological construct.
- **Construct excess**, where an existing modelling construct does not map to any existing ontological construct.
- **Construct deficit**, where an existing ontological construct does not map to any existing modelling construct.

The analysis of the ontology modelled using the ORM approach reveals that it suffers from both construct overload and construct deficit. In terms of overload, the 'book' class maps to books, book editions and copies. In terms of deficit, book editions and copies do not map to any existing modelling construct. These deficiencies are resolved via the re-engineering process, thus we consider that the OP ontology provides greater clarity in the resulting conceptualisation. The underlying importance of ontological clarity is that it affects human understanding of the represented phenomenon (Shanks et al. 2008).

Not only concepts, but also the properties (i.e., relationships) that exist within the ontology are examined throughout the process of re-engineering. To be more semantically precise, the re-engineering process links 'Titles', 'People', and 'ISBNs' to 'BookEditions' through the tuple types 'hasName', 'isWrittenBy' and 'hasIdentifier' respectively as these descriptors could change only if changes happen to editions (see Figure 4).

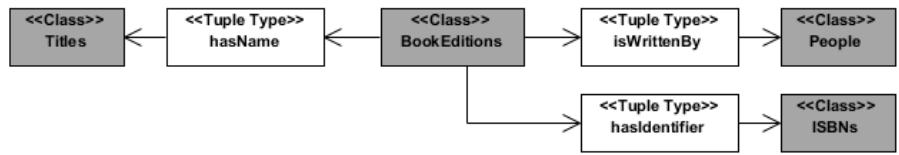

*Figure 4.The Re-engineered Properties relating to Book Editions*

The second set of competency questions is related to the ability of the ontology to track changes over time. The key issue here is the extent to which the ontology can capture changes happening to its objects, investigated via the following competency questions:

Q2.1 What is the price of a copy of a certain edition of a specific book at different points in time?

Q2.2 When has the price of a copy of a given edition of a specific book changed?

Q2.3 When has 'John Smith', for example, become an author?

As ORM is based on endurantism, where reality is modelled just as it is at a certain instant of time, the ORM ontology fails to answer the previous questions. To explain, let us consider the following example: My copy of 'Java how to program' was priced at £50 on 20th December 2005 and at £25 on 20th February 2009. So, the price of my copy was £50 between 20th December 2005 and 19th February 2009 and £25 from 20th February until now. With the previous analysis in mind, one can say that my copy of the book has two states which are created by two different events (two pricing assignment events).

As the OP is perdurantist it considers the temporal dimension, thus enables capturing of changes over time. Hence, we re-engineer the ontology and include a 'BookCopyStates' state-class along with a 'PriceAssignment' event-class in the OP ontology. Within the OP ontology, 'BookCopyStates' is linked to the 'Prices' class through the 'pricedAt' Tuple Type. In turn, the 'Price' class is linked to the 'Numbers' class and the 'Currencies' class by the 'hasValue' and 'hasUnits' Tuple Types respectively. In order to capture the time at which the price assignment has happened, the 'TimeInstants' class is connected to the 'PriceAssignment' event by the 'happensAt' tuple type (see Figure 5).

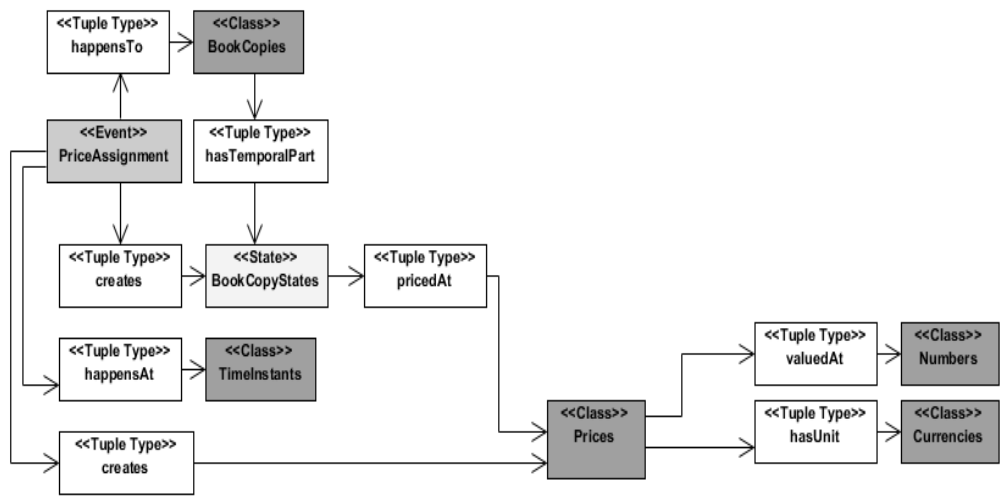

*Figure 5.The Re-engineered part relating to Book Price*

Such a re-engineering exercise allows us to capture more important details such as *record occurrence* and *record changes*, achieving a more natural and rigorous description of the phenomenon under investigation. The OP ontology is semantically richer and more precise as it is more capable of representing the real world accurately (see also Daga et al. 2005). This feature is highly important to modern information systems that are increasingly sophisticated and intelligent so as to respond to varied and more complex user requirements.

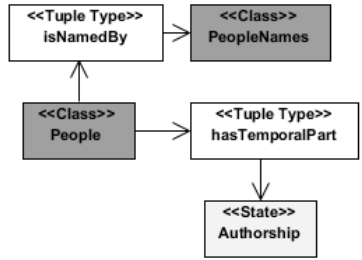

*Figure 6.The Re-engineered part relating to Authorship*

Finally, let us now move to analyse the 'Authorship' concept. In a four-dimensional paradigm, one can recognise that a person is actually an author during a specific period of their life. Being an author occupies just a stage of a person's life because they could have a different occupation, no occupation before being an author, or multiple occupations. Thus, in the OP ontology we model 'People' as a class where 'Authorship' is only one state of the People class. Therefore, the 'People' class is linked to the 'Authorship' state by the tuple type 'hasTemporalPart' (see Figure 6). On the other hand, people's names are obviously different objects from the people themselves who are called by these names. Therefore, in the OP ontology, the 'People' and 'PeopleNames' classes are linked by the 'isNamedBy' tuple type (see Figure 6).

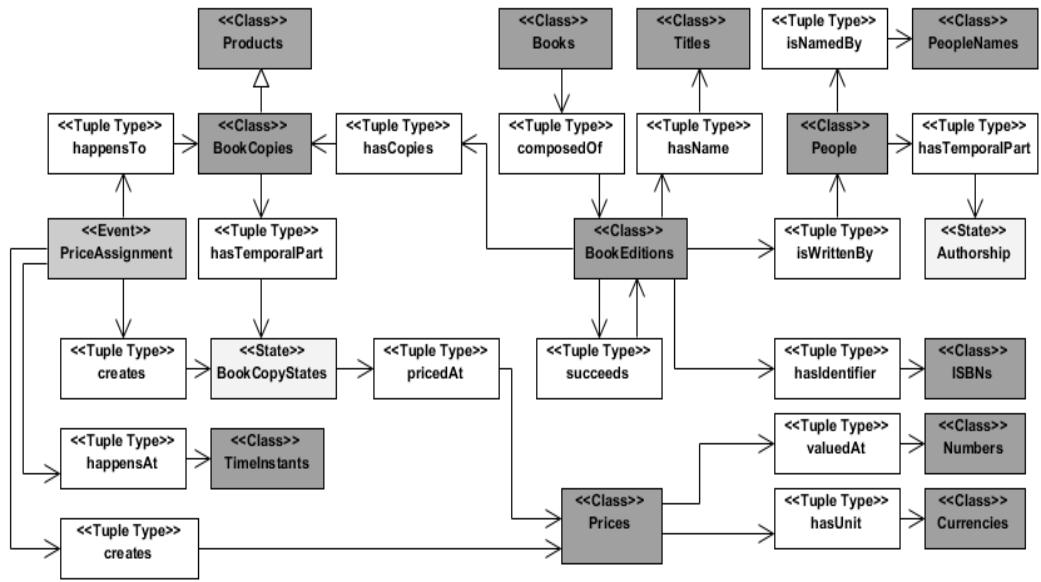

*Figure 7. The OP Bookstore ontology*

The re-engineered ontology is illustrated in Figure 7. Although ORM and OP have been both used in modelling ontologies, we propose that our analysis demonstrates that OP is more effective in this context. This is because, as Chandersekaran et al. (1999) argue, the following issues are important to be considered when conceptualising ontologies: (1) Real world objects and their properties change over time; (2) objects can go through different states which form temporal parts of these objects; and (3) states are created and dissolved by events. Capturing the previous important features is only possible by utilising OP due to the effective inclusion of time as a fourth dimension at the outset and not an afterthought.

# 5. DISCUSSION

In this section, we discuss four main aspects that are stimulated from the previous illustrative scenario. These aspects can be highlighted using the following questions:

1- Can endurantism approaches represent temporality? If not, what kind of modifications is required to do so?

2- Can an ontology language like the Web Ontology Language (OWL), for example, encodes ontologies conceptualised using OP?

3- May preferences of ontology engineers toward conceptualisation approaches differ across different application contexts?

4- What are the shortcomings and limitations of perdurantist approaches such as OP?

The aforementioned issues are considered important to the current research as they are meant to provide additional clarity to the differences between endurantism and perdurantism. They are also significant in showing the limitations of both endurantist and perdurantist approaches and hence the kind of future research that is needed so as to improve their use and effectiveness in practice. As for the first issue, although endurantist approaches treat time and space totally separately (Farrar and Bateman, 2004), one can argue that temporal aspects such as points in time as well as periods can be modelled using endurantism (for instance ER or OO) by adding an Entity Type or a Class named 'Time'. Thereafter, this Entity Type can be connected to other concepts, thus being able to capture temporal aspects. This argument is imprecise as it does not distinguish between conceptualisation and implementation phases during ontology engineering. Attaching a "Time" class would add very little to the conceptualisation produced using ER or OO. For example, by referring to our illustrative case presented in the previous section, this step would not allow us to capture price assignment as an event; neither would it allow us to consider 'BookCopyStates' as states during the conceptualisation phase. At this phase, the 'Time' class adds minimal detail to the conceptual model. It is only after the implementation phase, when we would be able to understand some of these details in ER or OO (i.e., events). This is because instances of classes would be stored in the ontology and linked to the 'Time' entity type only after the implementation phase. But even after the implementation, the details related to states of objects would not be able to be captured and represented explicitly since there would be no dedicated classes to represents states of different classes.

Another major deficiency of endurantist approaches in representing temporal aspects is related to the identity of objects. Even with the inclusion of 'Time' as an entity type, the identity of an object can be lost. Consider this example; an employee of a publisher was an author prior to being recruited by the publisher. The employed ontology has a class to capture details about the publisher's employees and another about authors' details where the primary key for the first class is employee number, whilst it is author number for 'Authors' class. Given this common example, it would be hard to pinpoint that a certain author is the same person who is an employee of the publisher now since they have different primary keys. In OP, on the other hand, as a representative of perdurantist approaches, identity of objects is maintained as all different states of any objects would be directly linked to the object itself and thus the identity is more likely to be kept intact.

Furthermore, and given that the philosophy of endurantist-based approaches assumes that objects are entirely present at each moment of their life time, it would be difficult for its adopters to be able to differentiate between books, book editions and book copies. Also, it would be not easy for them to perceive authorship as a temporal part of an individual person. For endurantist approaches to be able to capture temporal aspects effectively, we believe that two main modifications or extensions are required. The first modification is deemed core as it touches upon the approach's philosophy. To truly represent temporality, the approach needs to take into consideration that real-life objects and phenomena are dynamic as they go through various states in their lifetimes and that these states are controlled by events. After this key modification, endurantist approaches must modify and extend their modelling constructs in accordance with the new philosophy. Such modification, if it occurred, would create new perdurantist approaches or at least new versions of the original endurantist approaches that are able to deal effectively with temporality.

As for the second issue, modelling languages like OWL make no assumption as to the nature of the things that are modelled, i.e., whether these things are 3D or 4D for example. The same occurs with conventional data modelling techniques like ER models and ORM. In order to adopt a perdurantist approach, such as the Object Paradigm, it is necessary to construct a foundational ontology in which the core elements of the 4D ontology are represented. In the case of the object paradigm it would be necessary, as a minimal initial starting point, to represent core objects, without which no model could be produced, and a temporal ontology to manage the temporal relationships (between objects and their parts) and the fundamental types of temporal parts (states and events). The two parts of this foundational ontology would include therefore:

(1) Classes, Individuals, Tuples (individual relationships between objects) and Tuple Classes (classes of relationships).

(2) Types of temporal parts (states and events), types of relationship between 4D individuals and their 4D parts, time itself (as individual time instants and time intervals) and relationships between events/states and such time instants/periods.

The core part of the foundational ontology described above is represented in Figure 8.

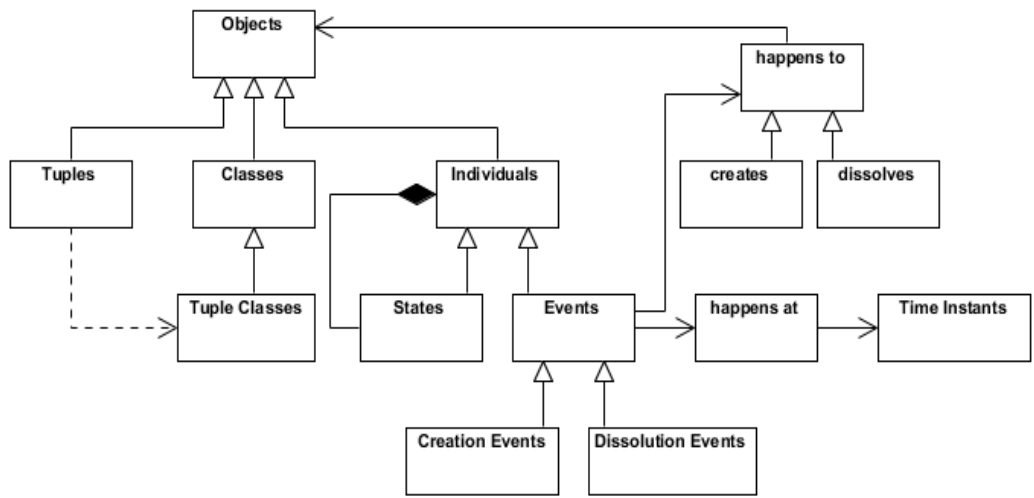

*Figure 8. The Upper Object Paradigm (OP) Ontology*

Obviously in OWL, as with any other modelling language, such classes can be represented, however one must keep in mind that people interpreting such models must always do so on the basis of the foundational ontology just mentioned. However, if OWL (or even another related Semantic Web language like RDF) is adopted, it would be more likely that this form of representation is required for machine processability. In this case any software that processes such 4D models must be designed to be "aware" of the perdurantist nature of the ontologies being used. Any reasoning that occurs on such models must incorporate rules that would apply to a 4D perspective of the world.

The third issue concerns whether preferences of ontology engineers toward conceptualisation approaches (ORM vs. OP) differ across different application contexts. Our view is (as stated earlier) that techniques like ORM make no assumption as to whether the world is modelled in a 3D or 4D way. Conventionally a 3D approach is adopted by knowledge, database and software designers; however such techniques can be applied to model a 4D universe if appropriate foundational ontologies are adopted and if the modeller makes strict use of such a foundation.

Moreover, even assuming that in the development of one specific software system it may be easier to adopt a 3D view, or that the existing requirements do not require a 4D representation to be satisfied, even in this situation the case can be made in favour of a 4D approach. In fact software development is mainly about evolution and change; hence software must be developed keeping in mind that it will undergo change (even substantial and fundamental change) over time. 4D approaches, such as the Object Paradigm, enable greater flexibility and extensibility in software.

As for the last issue which concerns the shortcomings of endurantist approaches, we believe that despite the ability of perdurantist modelling approaches, such as OP, in producing more semantically rich and faithful representations of real world phenomenon, their success and widespread use in practice are limited by some critical factors. For example, unlike ORM, the absence of well-established and unified graphical notations is one of the main shortcomings of OP as a conceptual modelling approach. Other shortcomings in OP are inherited from the first one. Indeed, without having such graphical notations, there will be no existing conceptual modelling software tools through which analysts and developers can easily create OP models and also translate such models into natural language statements, normalised database schemas, and into other useful forms such as markup language schemas.

The scarcity of research in information systems tackling and discussing issues related to endurantist conceptual modelling approaches is another key limitation. Given that this important area of research is underdeveloped, its high potential for improving conceptual modelling is somehow unseen. More research in this area would lead to a better understanding of such approaches; and thus better understanding of the additional benefits that could be achieved in the domain of conceptual modelling. Thereafter, research could focus on building graphical notations for such paradigms along with their software tools.

In the next section, we highlight the major implications for theory and practice generated by this research.

# 6. IMPLICATIONS FOR THEORY AND PRACTICE

The comparative analysis of endurantism and perdurantism using ORM and OP as representative paradigms shows important differences between current ontological modelling approaches. This research reveals that adopting different paradigms in modelling ontologies impacts the semantic quality of the resulting ontologies. The study also highlights that the OP is more effective as it enhances the semantic representation of real world phenomena. Such enhancements may also significantly affect the quality and performance of the implemented software systems. With hindsight, the theoretical and practical implications of this study can be summarised as follows:

- **Expressiveness and faithfulness.** Expressiveness is defined as the ability of a modelling paradigm to capture all conceptually relevant domain details in a clear manner (Haplin and Bloesch 1999). This feature is very important because a highly expressive paradigm is a useful tool that enables a modeller to conceive more details than what can be captured using a modestly expressible paradigm (reflecting these details appropriately in the produced model). The conducted re-engineering exercise shows that the OP provides more expressiveness compared to the ORM as more details are picked up during the re-engineering activities. Expressiveness is a very important feature of ontologies as it enables an ontology to provide a more faithful representation of the phenomenon it abstracts. Furthermore, as we discussed in the previous section, the re-engineered OP ontology resolves the problems related to construct overload and deficit which are present in the ORM ontology, allowing for a more clear representation of the phenomenon under investigation. Using endurantist approaches indicates moving from conceptual modelling to semantic conceptual modelling.

- **Temporality of produced models.** Information systems are becoming more complex and requiring continuous tracking to the occurrences and changes of things at different time frames in order to encapsulate more intelligence. To this aim, the ontology should be able to capture things in the past, present and potential future. This requirement cannot be modelled effectively using endurantist paradigms as they just model a phenomenon as it is at a specific time from the perspective of enduring entities. A model developed based on perdurantism allows the representation of things along with their changes (states and events) in a natural and precise manner. For example, modelling `BookCopyStates' as being temporal parts of `BookCopies' enables capturing the effect of price changes in the past, present and future. Another example is the `Authorship' state of `People'. This way of modelling reality is more accurate as people move through time from one occupation or stage to another. For instance, someone might be a student in one stage before becoming an author in a later one.

- **Extensibility to capture future business needs.** Extensibility is the ability of a model to respond to changes smoothly. In other words, a model should cater for new changes and needs without a substantial change occurring to its constructs. In a turbulent and dynamic environment, this issue has become well known and significant in the information systems arena. Adopting a perdurantist approach, however, one can produce a model that better responds to changes because of the inclusion of the temporal dimension. Any change happening to an object can be represented via states and events. To elaborate more, let us consider that a new business need has arisen requiring the ontology to capture the status (e.g., new or used) of book copies. To address this feature in the ORM ontology, a new role that could be called 'status' has to be added which might affect the unique identifier and necessitates some changes that may affect existing links to other components in the ontology-driven application. In the OP ontology, two states (new state and old state) can be added as being temporal parts of 'BookCopies'. This addition can be easily included in the existing ontology and will not affect any application using this ontology since these states are only temporal parts of an already existing object. Hence, such a significant feature leads to more flexible and reusable ontologies.

- **Objectivity.** A modelling process is arguably more objective when it provides clear guidelines for a modeller to eliminate confusion that might arise during the conceptualisation activity. Analysing the four dimensions of objects based on the BORO process (see Figure 1) distinguishes amongst the different objects and alleviates modelling errors and bias. For example, analysing the four dimensions of 'Books' leads us to add 'BookEditions' to the OP ontology as the four dimensions of a book are composed of the extensions of all the 'BookEditions' of this book.

Although the OP supports more semantically faithful representations of real world phenomena than ORM, its success is dependent on some factors such as: (1) The availability of more research on the OP; (2) the existence of supporting development tools; and (3) the ability of systems analysts and designers to understand and believe in the underlying philosophy behind such a paradigm.

# 7. CONCLUSIONS

Ontology is a relatively new innovation that promises to improve the design, semantic integration and utilisation of information systems. Ontologies increasingly provide the backbone of knowledge-based systems and are used to establish sharable and reusable understanding of specific domains amongst people, information systems and software agents. Notwithstanding, the ontology-related literature does not provide adequate guidance on how adopting different conceptual modelling paradigms during the analysis and design phase would impact the semantic quality of the developed ontologies and, perhaps, consequently the implemented software systems. To address this dilemma, this paper has analytically compared endurantist and perdurantist approaches using Object Role Modelling (ORM) and the Object Paradigm (OP) respectively as representatives.

The applied analysis reveals that OP (perdurantist) provides a semantically richer representation of the phenomena under investigation. The primary advantages are summarised as follows:

- *Better expressiveness*, providing clearer and more precise representation of reality based on spatial and temporal dimensions of objects.
- *Temporality*, providing the ability to capture dynamic objects showing their changes over time.
- *Better flexibility and reuse*, providing a more effective way of absorbing the changing business needs in the modern business environment.
- *Objectivity*, modelling real-life objects using a more systematic method that enjoys the merit of clearly mapping these objects into appropriate modelling constructs.

We acknowledge that the paper has some limitations in that it analyses the differences between endurantism and perdurantism by using only the ORM and OP approaches. Although the presented analysis shows some significant differences, still there is a need to delineate and validate such differences using other conceptual modelling approaches and this is the next stage of our research.

# References

[1] Al Asswad, M.M., de Cesare, S. and Lycett, M. (2009). Toward a Research Agenda for Semi-Automatic Annotation of Web Services, *International Conference on Informatics and Semiotics in Organisations (ICISO)* - IFIP WG8.1 Working Conference.

[2] Al-Debei, M. M. and Fitzgerald, G. (2009). OntoEng: A Design Method for Ontology Engineering in Information Systems. *ACM SIGPLAN International Conference on Object Oriented Programming, Systems, Languages and Applications*, ODiSE Workshop, Orlando, Florida, , 1-25.

[3] Allen, E., Edwards, G., and Bedard, Y. (1995). Qualitative Casual Modelling in Temporal GIS. *In Proceedings of the International Conference on Spatial Information Theory*, 397-412.

[4] Chandrasekaran, B., Josephson, J.R. and Benjamins, V.R. (1999). What are Ontologies, and Why Do We Need Them?. *IEEE Intelligent Systems*, 14 (1), 20-26.

[5] Daga, A., De Cesare, S., Lycett, M. and Partridge, C. (2005). An Ontological Approach for Recovering Legacy Business Content. *In Proceedings of the 38th Annual Hawaii International Conference onSystem Science*, 1-9.

[6]     Devedzic, V. (2002). Understanding ontological engineering. *Communications of the ACM*. 45 (4), 136-144.

[7]     Erwig, M., Guting, R.H., Schneider, M. and Vazirgiannis, M. (1999). Spatio-temporal data types: an approach to modeling and querying moving objects in databases. *GeoInformatica*, 3 (3), 269-296.

[8]     Fernández-López, M., Gómez-Pérez, A., Sierra, J.P. and Sierra, A.P. (1999). Building a Chemical Ontology Using Methontology and the Ontology Design Environment. *IEEE Intelligent Systems*, 14 (1), 37-46.

[9]     Fonseca, F. (2007). Learning the Differences Between Ontologies and Conceptual Schemas Through Ontology-Driven Information Systems., *JAIS - Journal of the Association for Information Systems - Special Issue on Ontologies in the Context of IS*, 8 (2), 129-142.

[10]   Gómez-Pérez, A., Corcho, O. and Fernández-López, M. (2004). 'Ontological Engineering: with examples from the areas of Knowledge Management, e-Commerce and the Semantic Web'. Fifth edn, Springer.

[11]   Gómez-Pérez, A. (2001). Evaluation of Ontologies. *International Journal of Intelligent Systems*, 16 (3), 391-409.

[12]   Gruber, T.R. (1993). A Translation Approach to Portable Ontology Specification. *Knowledge Acquisition*, 5 (2), 199-220.

[13]   Gruber, T.R. (1995). Toward principles for the design of ontologies used for knowledge sharing. *International Journal of Human-Computer Studies*, 43 (5-6), 907-928.

[14]   Gruninger, M. and Fox, M.S. (1995). Methodology for the design and evaluation of ontologies. *In Proceedings of the Workshop on Basic Ontological Issues in Knowledge Sharing*, 1-10.

[15]   Guarino, N. (1998). Formal Ontology in Information Systems. *In Proceedings of Formal Ontology in Information Systems Conference 98*, 3-15.

[16]   Guarino, N. and Welty, C. (2002). Evaluating ontological decisions with OntoClean. *Communications of the ACM*, 45 (2), 61-65.

[17]   Guarino, N. (2004). Toward a Formal Evaluation of Ontology Quality. *IEEE intelligent Systems*, 19 (4), 78-79.

[18]   Hadzilacos, T. and Tryfona, N. (1997). Extending the Entity-Relational Model to Capture Spatial Semantics. *SIGMOD Record*, 26 (3).

[19]   Hales, S.D. and Johnson, T.A. (2003). Endurantism, Perdurantism and Special Relativity.*The Philosophical Quarterly*, 53 (213), 524-539.

[20]   Halpin, T. (1998). Object-Role Modeling (ORM/NIAM). *In Handbook on Architectures of Information Systems*, 1-9.

[21]   Halpin, T. and Bloesch, A. (1999). Data modeling in UML and ORM: a comparison. *Journal of Database Management*, 10 (4), 4-13.

[22]   Jarrar, M., Demey, J. and Meersman, R. (2003). On Using Conceptual Data Modeling for Ontology Engineering. *Journal on Data Semantics,* 1(1), LNCS 2008, 185-207.

[23]   Jasper, R. and Uschold, M. (1999). A Framework for Understanding and Classifying Ontology Applications. *Workshop on Ontologies and Problem-Solving Methods*.

[24]   Kishore, R., R. Sharman, and R. Ramesh (2004) "Computational Ontologies and Information Systems I: Foundations," *Communications of the Association for Information Systems* (14), 158-183.

[25] Krogstie, J., Opdahl, A.L. and Brinkkemper, S. (2007). 'Conceptual Modeling in Information Systems Engineering', Springer.

[26] Moro, M.M., Saggiorato, S.M., Edelweiss, N. and Santos, C.S. (2001). Adding Time to an Object-Oreinted Versions Model. *In Proceedings of the 12th International Conference on Databases and Expert Systems Applications*, Springer-Verlag, 805-814.

[27] Mylopoulos, J. (1992). 'Conceptual modeling and Tools'. In Conceptual Modeling, Databases and CASE, eds. P. Loucopoulos & R. Zicari, Wiley, 49-68.

[28] Nicola, A.D., Missikoff, M. and Navigli, R. (2009). A software engineering approach to ontology building. *Information Systems*, 34 (2), 258-275.

[29] Ouksel, A.M. and Sheth, A. (1999). Semantic interoperability in global information systems. *ACM SIGMOD Record*, 28 (1), 5-12.

[30] Parent, C., Spaccapietra, S. and Zimanyi, E. (1999). Spatio-temporal conceptual models: data structure + space + time. *In Proceedings of the 7th ACM international symposium on Advances in geographic information systems ACM*, New York, USA, 26-33.

[31] Partridge, C. (2001). The BORO Approach, BORO Centre. Available at: (http://www.boroprogram.org/boro_centre/ , accessed on: 28th of June 2009).

[32] Partridge, C. (2005). 'Business Objects Re-Engineering for Re-Use'. 2nd edn, The BORO Centre, London.

[33] Pinto, H.S. and Martins, J.P. (2004). Ontologies: How can they be Built?. *Knowledge and Information Systems*, 6 (4), 441-464.

[34] Shanks, G., Tansley, E., Nuredini, J., Tobin, D. and Weber, R. (2008). Representing Part-Whole Relations in Conceptual Modeling: An Empirical Evaluation. *MIS Quarterly*, 32(3), 553-573.

[35] Spyns, P., Meersman, R. and Jarrar, M. (2002). Data modeling versus Ontology engineering. *ACM SIGMOD Record*, 31 (4), 12-17.

[36] Uschold, M. and King, M. (1995). Towards a Methodology for Building Ontologies. *Workshop on Basic Ontological Issues in Knowledge Sharing*.

[37] Wand, Y. and Weber, R. (1990). An ontological model of an information system. *IEEE Transactions on Software Engineering*, 16 (11), 1282-1292.

[38] Wand, Y. and Weber, R. (2002). Research Commentary: Information Systems and Conceptual Modeling--A Research Agenda. *Information Systems Research*, 13 (4), 363-376.

[39] Worboys, M. (1994). A Unified Model for Spatial and Temporal Information. *The Computer Journal*, 37 (1), 26-34.

[40] Xu, W., Huang, H.K. and Liu, X.H. (2006). Spatio-Temporal Ontology and its Application in Geographic Information System. *In Proceedings of the Fifth International Conference on Machine Learning and Cybernetics*, 1487-1492.